\documentclass[accepted]{article}

\usepackage{microtype}
\usepackage{graphicx}
\usepackage{subfig}
\usepackage{booktabs} 

\usepackage{hyperref}

\usepackage{amsmath}
\usepackage{mathrsfs}
\usepackage[mathscr]{eucal}
\usepackage{amssymb}
\usepackage{authblk}
\usepackage{color}
\usepackage{comment, float}
\usepackage{mathptmx} 
\usepackage{bm}
\usepackage{subfloat}
\usepackage{icml2020}

\usepackage{setspace}

\icmltitlerunning{}

\begin{document}

\twocolumn[
\icmltitle{Incorporating Relational Background Knowledge into Reinforcement Learning via Differentiable Inductive Logic Programming}



\icmlsetsymbol{equal}{*}

\begin{icmlauthorlist}
\icmlauthor{Ali Payani}{ga}
\icmlauthor{Faramarz Fekri}{ga}
\end{icmlauthorlist}

\icmlcorrespondingauthor{Ali Payani}{payani@ece.gatech.edu}

\icmlaffiliation{ga}{Department of Electrical an Computer Engineering, Georgia Institute of Technology}
\icmlkeywords{Machine Learning, ICML}
\vskip 0.3in
]



\printAffiliationsAndNotice{\icmlEqualContribution} 


\begin{abstract}
Relational Reinforcement Learning (RRL) can offers various desirable features. Most importantly, it allows for incorporating expert knowledge into the learning, and hence leading to much faster learning and better generalization compared to the standard deep reinforcement learning. However, most of the existing RRL approaches are either incapable of incorporating expert background knowledge (e.g., in the form of explicit predicate language) or are not able to learn directly from non-relational data such as image. 
In this paper, we propose a novel deep RRL based on a differentiable Inductive Logic Programming (ILP) that can effectively learn relational information from image and present the state of the environment as first order logic  predicates. Additionally, it can take the expert background knowledge and incorporate it into the learning problem using appropriate predicates. The differentiable ILP allows an end to end optimization of the entire framework for learning the policy in RRL. We show the efficacy of this novel RRL framework using environments such as BoxWorld, GridWorld as well as relational reasoning for the Sort-of-CLEVR dataset.
\end{abstract}

\section{Introduction}
\label{sec:INTRO}
Relational Reinforcement Learning (RRL) has been investigated in early 2000s by works such as \cite{bryant1999combining,dvzeroski1998relational,dvzeroski2001relational} among others. 
The main idea behind RRL is to describe the environment in terms of objects and relations. 
One of the first practical implementation of this idea was proposed by \cite{dvzeroski1998relational} and later was improved in \cite{dvzeroski2001relational} based on a modification to Q-Learning algorithm \cite{watkins1992q} via the standard relational tree-learning algorithm TILDE \cite{blockeel1998top}. As shown in \cite{dvzeroski2001relational}, the RRL system allows for  very natural and human readable decision making and policy evaluations. More importantly, the use of variables in ILP system, makes it possible to learn generally formed policies and strategies. Since these policies and actions  are not directly associated with any particular instance and entity, this approach leads to a generalization capability beyond what is possible in most typical RL systems.
Generally speaking RRL framework offers several benefits over the traditional RL: (i) The learned policy is usually human interpretable, and hence can be viewed, verified and even tweaked by an expert observer. (ii) The learned program can generalize better than the classical RL counterpart. (iii) Since the language for the state representation is chosen by the expert, it is possible to incorporate inductive biases into learning. This can be a significant improvement in complex problems as it might be used to manipulate the agent to choose certain actions without accessing the reward function, (iv) It allows for the incorporation of higher level concepts and prior background knowledge. 

In recent years and with the advent of the new deep learning techniques, significant progress has been made to the classical Q-learning RL framework. By using algorithms such as deep Q-learning and its variants \cite{mnih2013playing,van2016deep}, as well as Policy learning algorithms such A2C and A3C \cite{mnih2016asynchronous}, more complex problems are now been tackled. However, the classical RRL framework cannot be easily employed to tackle  large scale and complex scenes that exist in recent RL problems. 
Since standard RRL framework is in not usually able to learn from complex visual scenes and cannot be easily combined with differentiable deep neural 
In particular, none of the inherent benefits of RRL have been materialized in the deep learning frameworks thus far. This is because existing RRL frameworks usually are not designed to learn from complex visual scenes and cannot be easily combined with differentiable deep neural networks.
In ~\cite{payani2019Learning} a novel ILP solver was introduced which uses Neural-Logical Network (NLN)~\cite{payani2018} for constructing a differentiable neural-logic ILP solver (dNL-ILP). The key aspect of this dNL-ILP solver is a differentiable deduction engine which is at the core of the proposed RRL framework. 
As such, the resulting differentiable RRL framework can be used similar to deep RL in an end-to-end learning paradigm, trainable via the typical gradient optimizers. Further, in contrast to the early RRL frameworks, this framework is flexible and can learn from ambiguous and fuzzy information. Finally, it can be combined with deep learning techniques such as CNNs to extract relational information from the visual scenes. 
In the next section we briefly introduce the differentiable dNL-ILP solver. In section, \ref{sec:RRL} we show how this framework can be used to design a differentiable RRL framework. Experiments will be presented next, followed by the conclusion.

\section{Differentiable ILP via neural logic networks}
\label{sec:dNL-ILP}
 
\begin{figure}
	\centering
	\includegraphics[width=0.2\textwidth]{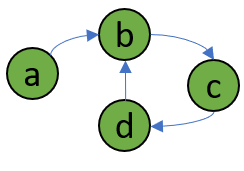}
	\caption{Connected graph example}
	\label{fig:connected_graph}
	\vspace{-.16in}
\end{figure}

In this section, we briefly present the basic design of the differentiable dNL-ILP which is at the core of the proposed RRL. More detailed presentation of dNL-ILP could be found in \cite{payani2019Learning}.
Logic programming is a paradigm in which we use formal logic (and usually first-order-logic) to describe relations between facts and rules of a program domain. In logic programming, rules are usually written as clauses of the form $H \leftarrow B_1,\,B_2,\,\dots,\,B_m$, 
where $H$ is called \texttt{head} of the clause and $B_1,\,B_2,\,\dots,\,B_m$ is called \texttt{body} of the clause. A clause of this form expresses that if all the atoms $B_i$ in the \texttt{body} are true, the \texttt{head} is necessarily true.
Each of the terms $H$ and $B$ is made of \texttt{atoms}. Each \texttt{atom} is created by applying an $n$-ary Boolean function called \texttt{predicate} to some constants or variables. A \texttt{predicate} states the relation between some variables or constants in the logic program. We use lowercase letters for constants (instances) and uppercase letters for variables. 
To avoid technical details, we consider a simple logic program. Assume that a directed graph is defined using a series of facts in the form of  \texttt{edge(X,Y)} where for example \texttt{edge(a,b)} states that there is an edge from node \texttt{a} to the node \texttt{b}. As an example, the graph in Fig. \ref{fig:connected_graph} can be represented as \texttt{\{edge(a,b), edge(b,c), edge(c,d), edge(d,b)\}}.
Assume that our task is to learn the \texttt{cnt(X,Y)} predicate from a series of examples, where \texttt{cnt(X,Y)} is true if there is a directed path from node \texttt{X} to node \texttt{Y}. The set of positive examples in graph depicted in Fig. \ref{fig:connected_graph} is $\mathcal{P}=$ \texttt{\{cnt(a,b), cnt(a,c), cnt(a,d), cnt(b,b),cnt(b,c), cnt(b,d),\dots\}}. Similarly the set of negative examples  $\mathcal{N}$ includes atoms such as \texttt{\{cnt(a,a),cnt(b,a),\dots\}}.

It is easy to verify that the predicate \texttt{cnt} defined as below satisfies all the provided examples (entails the positive examples and rejects the negative one):
\begin{align}
\text{cnt(X,Y)} &\leftarrow \text{edge(X,Y)} \nonumber\\
\text{cnt(X,Y)} &\leftarrow \text{edge(X,Z),\,\,cnt(Z,Y)}
\label{eq:cnt}
\end{align}
In fact by applying each of the above two rules to the constants in the program we can produce all the consequences of such hypothesis
If we allow for formulas with 3 variables (\texttt{X,Y,Z}) as in (\ref{eq:cnt}), we can easily enumerate all the possible symbolic atoms that could be used in the body of each clause. In our working example, this corresponds to $\mathbb{I}_{cnt}=$\texttt{\{edge(X,X), edge(X,Y), edge(X,Z), \dots, cnt(Z,Y), cnt(Z,Z)\}}. 
%
As the size of the problem grows, considering all the possibilities becomes unfeasible. Consequently, almost all ILP systems use some form of rule templates to reduce the possible combinations. For example, the dILP \cite{evans2018learning} model, allows for the clauses (in the body) of at most two atoms in each clause predicate. 
In \cite{payani2019Learning}, a novel approach was introduced to alleviate the above limitation and to allow for learning arbitrary complex predicate formulas. The main idea behind this approach is to use multiplicative neurons \cite{payani2018} that are capable of learning and representing Boolean logic. Consider the fuzzy notion of Boolean algebra where fuzzy Boolean value are represented as a real value in range $[0,1]$, where True and False are represented by 1 and 0, respectively. Let $\bar{x}$ be the logical `NOT' of $x$. 
%
%
\begin{figure}[tb]
	\centering
	\subfloat[][]{
		\small
		\vspace{-5mm}
		\begin{tabular}{|c|c|c|}
			\hline  
			$x_i$ & $m_i$ & $F_c$ \\ 	\toprule  
			0 & 0 & 1 \\ \hline
			0 & 1 & 0 \\ \hline
			1 & 0 & 1 \\ \hline
			1 & 1 & 1 \\ \hline
		\end{tabular}
		\label{fig:Fc}%
	}%
	\qquad
	\subfloat[][]{
		\small
		\begin{tabular}{|c|c|c|}
			\hline	 
			$x_i$ & $m_i$ & $F_d$ \\   	\toprule
			0 & 0 & 0 \\ \hline
			0 & 1 & 0 \\ \hline
			1 & 0 & 0 \\ \hline
			1 & 1 & 1 \\ \hline
		\end{tabular}
		\label{fig:Fd}%
	}
	\caption{Truth table of $F_c(\cdot)$ and $F_d(\cdot)$ functions}%
	\label{fig:FcFd}%
\end{figure}
Let $\boldsymbol{x}^n \in \{0,1\}^n$ be the input vector for a logical neuron. we can associate a trainable Boolean membership weight $m_i$ to each input elements $x_i$ from vector $\boldsymbol{x}^n$. Consider Boolean function $F_c(x_i,m_i)$ with the truth table as in Fig. \ref{fig:Fc} which is able to include (exclude) each element  $x_i$ in (out of) the conjunction function $f_{conj}(\boldsymbol{x}^n)$. This design ensures the incorporation of each element $x_i$ in the conjunction function only when the corresponding membership weight $m_i$ is $1$. Consequently, the  neural conjunction function $f_{conj}$ can be defined as:
\vspace{-2mm}
\begin{align}
\label{eq:conj}
f_{conj}(\boldsymbol{x}^n) &= \prod_{i=1}^{n} F_c(x_i,m_i) \nonumber \\
\text{where, }  \quad F_c(x_i,m_i) &= \overline{\overline{x_i} m_i } = 1 - m_i ( 1 - x_i) 
\end{align}
Likewise, a neural disjunction function $f_{disj}(\boldsymbol{x}^n) $  can be defined using the auxiliary function $F_d$ with the truth table as in Fig. ~\ref{fig:Fd}. 
By cascading a layer of $N$ neural conjunction functions with a layer of $N$ neural disjunction functions, we can construct a differentiable function to be used for representing and learning a Boolean Disjunctive Normal Form (DNF). 
%

dNL-ILP employs these differentiable Boolean functions (e.g. dNL-DNF) to represent and learn predicate functions. Each dNL function can be seen as a parameterized symbolic formula where the (fuzzy) contribution of each symbol (atom) in the learned hypothesis is controlled by the trainable membership weights (e.g., $w_i$ where $m_i = sigmoid(w_i)$). If we start from the background facts ( e.g. all the groundings of predicate \texttt{edge(X,Y)} in the graph example and apply the parameterized hypothesis we arrive at some new consequences (e.g., forward chaining). After repeating this process to obtain all possible consequences, we can update the parameters in dNL by minimizing the  cross entropy between the desired outcome (provided positive and negative examples) and the deduced consequences. 

%
An ILP description of a problem in this framework consist of these elements:
\begin{enumerate}
	\item The set of constants in the program. In example of Fig. \ref{fig:connected_graph}, this consists of $\mathcal{C}=$\texttt{\{a,b,c,d\}}
	\item The set of background facts. In the graph example above this consists of groundings of predicate \texttt{edge(X,Y)}, i.e., $\mathcal{B}=$\texttt{\{edge(a,b), edge(b,c), edge(c,d), edge(d,b)\}}
	
	\item The definition of auxiliary predicates. In the simple example of graph we did not include any auxiliary predicates. However, in more complex example they would greatly reduce the complexity of the problem.
	
	\item The signature of the target hypothesis. In the graph example, This signature indicates the target hypothesis is 2-ary predicate \texttt{cnt(X,Y)} and in the symbolic representation of this Boolean function we are allowed to use three variables \texttt{X,Y,Z}. 
	
	
\end{enumerate}
In addition to the aforementioned elements, some parameters such as intial values for the membership weights ($m_i=sigmoid(w_i)$), as well as the number of steps of forward chaining should be provided. Furthermore, in dNL-ILP the memberships are fuzzy Boolean values between 0 and 1. As shown in \cite{payani2019Learning}, for ambiguous problems where a definite Boolean hypothesis may not be found which could satisfy all the examples, there is no guaranty that the membership weights converge to zero or 1. In applications where our only goal is to find a successful hypothesis this result is satisfactory. However, if the interpretability of the learned hypothesis is by itself a goal in learning, we may need to encourage the membership weights to converge to 0 and 1 by adding a penalty term:
\begin{equation}
\text{interpretability penalty} \propto m_i(1-m_i)\label{eq:interpret}
\end{equation}

%
%
%
%
\section{Relational Reinforcement Learning via dNL-ILP}
\label{sec:RRL}

Early works on RRL \cite{dvzeroski2001relational,van2005survey} mainly relied on access to the explicit representation of states and actions  in terms of relational predicate language. In the most successful instances of these approaches, a regression tree algorithm is usually used in combination with a  modified version of Q-Learning algorithms. 
The fundamental limitation of the traditional RRL approaches is that the employed ILP solvers are not differentiable. Therefore, those approaches are typically only applicable the problems for which the explicit relational representation of states and actions is provided. Alternatively, deep RL models, due to recent advancement in deep networks, have been successfully applied to the much more complex problems. These models are able to learn from raw images without relying on any access to the explicit representation of the scene. However, the existing RRL counterparts are falling behind such desirable developments in deep RL. 

In this paper, we establish that differentiable dNL-ILP provides a platform to combine RRL  with deep learning methods, constructing a new RRL framework with the best of both worlds. This new RRL system allows the model to learn from the complex visual information received from the environment and extract intermediate explicit relational representation from the raw images by using the typical deep learning models such as convolutional networks. 
\begin{figure*}
	\centering
	\includegraphics[width=0.85\textwidth]{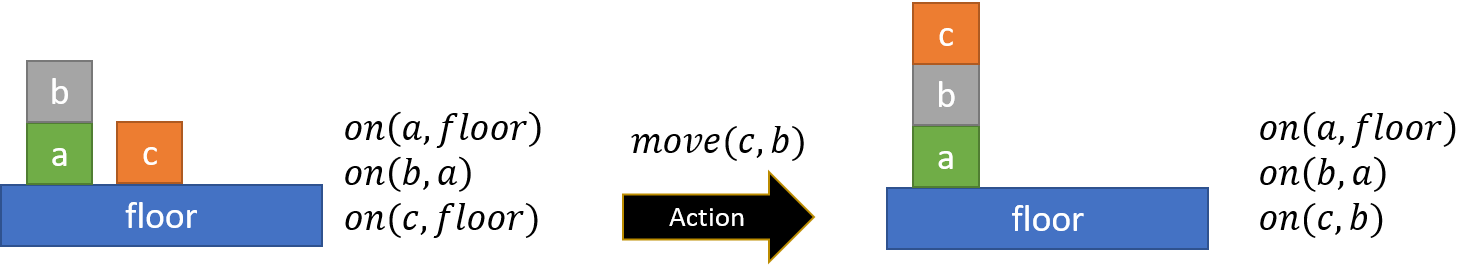}
	\caption{States representation in the form of predicates in BoxWorld game,  before and after an action}
	\label{fig:boxworld}
\end{figure*}
%
Although the dNL-ILP can also be used to formulate RL algorithms such as deep Q-learning, we focus only on deep policy gradient learning algorithm. This formulation is very desirable because it makes the learned policy to be interpretable by human. 
One of the other advantages of using policy gradient in our RRL framework is that it enables us to restrict actions according to some rules obtained either from human preferences or from problem requirements. This in turn makes it possible to account for human preferences or to avoid certain pitfalls, e.g.,  as in safe AI.

\begin{figure*}[h]
	\centering
	\includegraphics[width=.95\textwidth]{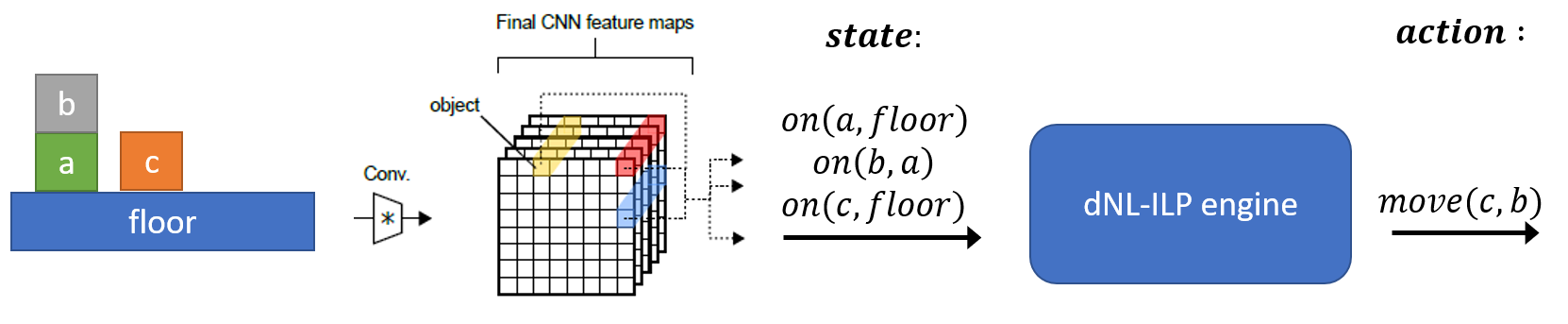}
	\vspace{-5mm}
	\caption{Learning explicit relational information from images in our proposed RRL; Images are processed to obtain explicit representation and dNL-ILP engine learns and expresses the desired policy (actions)}
	\label{fig:ilp_rrl_diag}
\end{figure*}

In our RRL framework, although we use the generic formulation of the policy gradient with the ability to learn stochastic policy, certain key aspects are different from the traditional deep policy gradient methods, namely state representation, language bias and action representation. In the following, we will explain these concepts in the context of BoxWorld game. In this game, the agent's task is to learn how to stack the boxes on top of each other (in a certain order). For illustration, consider the simplified version of the game as in Fig.\ref{fig:boxworld} where there are only three boxes labeled as \texttt{a,b}, and \texttt{c}. A box can be on top of another or on the \texttt{floor}. A box can be moved if it is not covered by another box and can be either placed on the floor or on top of another uncovered box. For this game, the environment state can be fully explained via the predicate \texttt{on(X,Y)}. Fig. \ref{fig:boxworld} shows the state representation of the scene before and after an  action (indicated by the predicate \texttt{move(c,b)}). In the following we discuss each distinct elements of the proposed framework using the BoxWorld environment. 
Fig.~\ref{fig:ilp_rrl_diag} displays the overall design of our proposed RRL framework. In the following we discuss the elements of this RRL system.

%
%
\begin{figure*}
	\centering
	\includegraphics[width=1\textwidth]{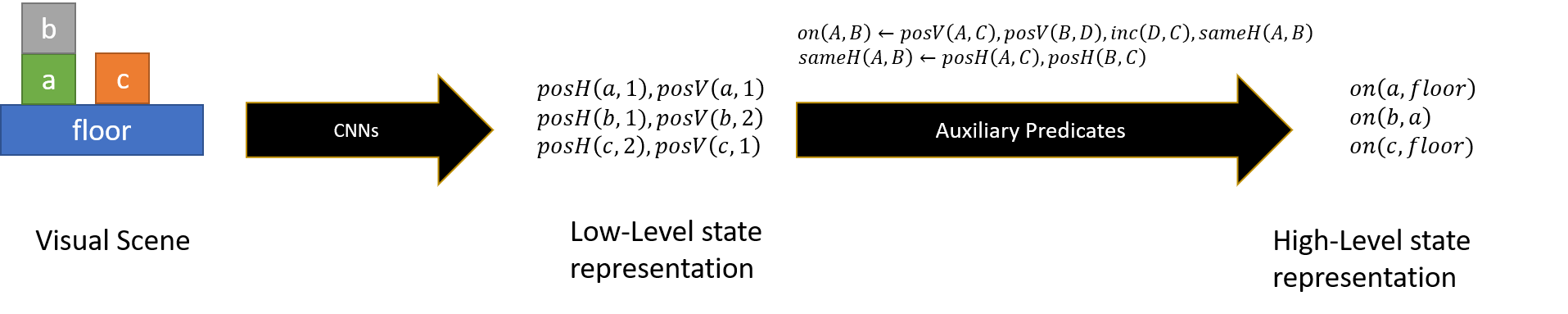}
	\vspace{-10mm}
	\caption{Transforming low-level state representation to high-level form via auxiliary predicates}
	\label{fig:ilp_rrl_state}
\end{figure*}
\begin{figure*} 
	\centering     
	\subfloat[A sample from CLEVER datset]{
		\includegraphics[width=.55\textwidth]{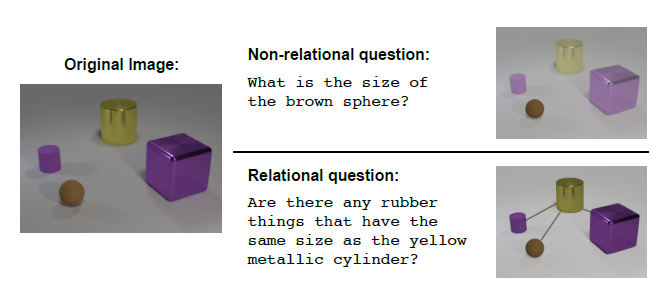}
	}
	\subfloat[A sample from sort-of-CLEVER datset]{
		\includegraphics[width=.28\textwidth]{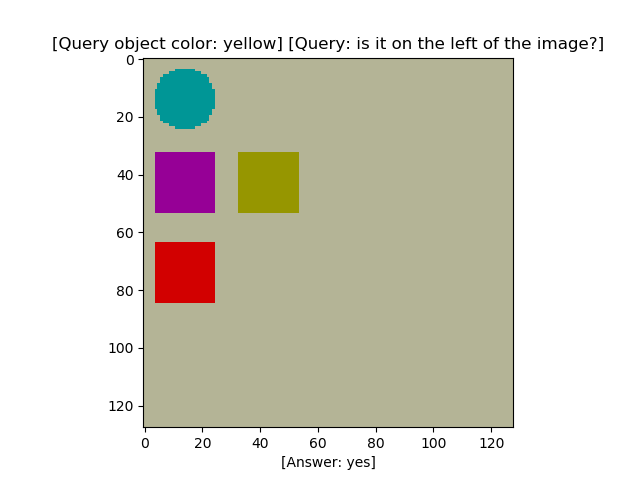}
		\label{subfig:sortofclever}
	}
	\vspace{-2mm}
	\caption{Extracting relational information from visual scene \cite{santoro2017simple}}
	\label{fig:ilp_RELATIONAL}
	
\end{figure*}
\subsection{State Representation}
\label{subsec:STATE-REP}
In the previous approaches to the RRL \cite{dvzeroski1998relational,dvzeroski2001relational,jiang2019neural}, state of the environment is expressed in an explicit relational format in the form of predicate logic. This significantly limits the applicability of RRL in complex environments where such representations are not available. Our goal in this section is to develop a method in which the explicit representation of states can be learned via typical deep learning techniques in a form that will support the policy learning via our differentiable dNL-ILP. 
As a result, we can utilize the various benefits of the RRL discipline without being restricted only to the environments with explicitly represented states.

For example, consider the BoxWorld environment explained earlier where the predicate
\texttt{on(X,Y)} is used to represent the state explicitly in the relational form (as shown in Fig.\ref{fig:boxworld}). 
Past works in RRL relied on access to explicit relational representation of states, i.e.,
all the groundings of the state representation predicates. 
Since this example has 4 constants, i.e. $\mathcal{C}=$\texttt{\{a,b,c,floor\}}, these groundings would be the binary values (`true' or `false') for the atoms \texttt{on(a,a), on(a,b), on(a,c), on(a,floor), \dots, on(floor,floor)}. 
%
%
In recent years, extracting relational information from visual scenes has been investigated.
Fig. \ref{fig:ilp_RELATIONAL} shows two types of relational representation extracted from images in~\cite{santoro2017simple}. 
The idea is to first process the images through multiple CNN networks. The last layer of the convolutional network chain is treated as the feature vector and is usually augmented with some non-local information such as absolute position of each point in the final layer of the CNN network. This feature map is then fed into a relational learning unit which is tasked with extracting non-local features.
Various techniques have been then introduced recently for learning these non-local information from the local feature maps, namely, self attention network models \cite{vaswani2017attention,santoro2017simple} as well as graph networks \cite{narayanan2017graph2vec,allamanis2017learning}. Unfortunately, none of the resulting presentations from past works is in the form of predicates needed in ILP.
%
%

In our approach, we use similar networks discussed earlier to extract non-local information. However given the relational nature of state representation in our RRL model, we consider three strategies in order to facilitate learning the desired relational state from images. Namely:
\begin{enumerate}
	\item \textbf{Finding a suitable state representation:} In our BoxWorld example, we used the \texttt{on(X,Y)} to represent the state of the environment. However, learning this predicate requires inferring relation among various objects in the scene. As shown by previous works (e.g., \cite{santoro2017simple}), this is a difficult task even in the context of a fully supervised setting (i.e., all the labels are provided) which is not applicable here. Alternatively, we propose to use lower-level relation for state representation and build higher level representation via predicate language. In the game of BoxWorld as an example, we can describe states by the respective position of each box. In particular, we  define two predicates \texttt{posH(X,Y)} and \texttt{posV(X,Y)} such that variable $X$ is associated with the individual box, whereas $Y$ indicate horizontal or vertical coordinates of the box, respectively. Fig.~ \ref{fig:ilp_rrl_state} shows how this new lower-level representations can be transformed into the higher level description by the appropriate predicate language: 
	\begin{align}
	\text{on(X, Y)} &\leftarrow \text{posH(X, Z)}, \text{posH(Y, T)}, \nonumber\\ &\text{inc(T, Z)}, \text{sameH(X, Y)} \nonumber\\
	\text{sameH(X, Y)} &\leftarrow \text{posH(X, Z)}, \text{posH(Y, Z)}
	\label{eq:on}
	\end{align}
	\item \textbf{State constraints:} When applicable, we may incorporate relational constraint in the form of a penalty term in the loss function. For example, in our BoxWorld example we can notice that \text{posY(floor)} should be always 0. In general, the choice of relational language makes it possible to pose constraints based on our knowledge regarding the scene. Enforcing these constraints does not necessarily speed up the learning as we will show in the BoxWorld experiment in Section \ref{subsec:BoxWorld}. However, it will ensure that the (learned) state representation and consequently the learned relational policy resemble our desired structure of the problem.
	\item \textbf{Semi-supervised setting:} While it is not desirable to label every single scene that may happen during learning, in most cases it is possible to provide a few labeled scene to help the model to learn the desired state representation faster. These reference points  can then be incorporated to the loss function to encourage the network to learn a representation that matches to those provided labeled scenes. We have used a similar approach in Asterix experiment (see appendix \ref{app:asterix}) to significantly increase the speed of learning.
\end{enumerate}

\subsection{Action Representation}
\label{subsec:ACTION-REP}
We formulate the policy gradient in a form that allows the learning of the actions via one (or multiple) target predicates. These predicates exploit the background facts, the state representation predicates, as well as auxiliary predicates to incorporate higher level concepts. 
In a typical Deep policy gradient (DPG) learning, the probability distributions of actions are usually learned by applying a multi layer perception with a \texttt{softmax} activation function in the last layer. In our proposed RRL, the action probability distributions can usually be directly associated with groundings of an appropriate predicate. For example, in  BoxWorld example in Fig.\ref{fig:boxworld}, we define a predicate \texttt{move(A,B)} and associate the actions of the agent with the groundings of this predicate. In an ideal case, where there is deterministic solution to the RRL problem, the predicate \texttt{move(A,B)} may be learned in such a way that, at each state, only the grounding (corresponding to the correct action) would result 1 ('true') and all the other groundings of this predicate become 0. In such a scenario, the agent will follow the learned logic deterministically. Alternatively, we may get more than one grounding with value equal to 1 or we get some fuzzy values in the range of $[0,1]$. 
In those cases, we estimate the probability distribution of actions similar to the standard deep policy learning by applying a \texttt{softmax} function to the valuation vector of the learned predicate \texttt{move} (i.e., the value of \texttt{move(X,Y)} for \texttt{X,Y}$\in$ \texttt{\{a,b,c,floor\}}).  
%
%

\section{Experiments}
\label{sec:EXPERIMENTS}
In this section we explore the features of the proposed RRL framework via several examples. We have implemented\footnote{The python implementation of the algorithms in this paper
	is available at \url{https://github.com/dnlRRL2020/RRL}} the models using Tensorflow \cite{abadi2016tensorflow}.
\subsection{BoxWorld Experiment}
\label{subsec:BoxWorld}
BoxWorld environment has been widely used as a benchmark in past RRL systems \cite{dvzeroski2001relational,van2005survey,jiang2019neural}. In these systems the state of the environment is usually given as an explicitly relational data via groundings of the predicate \texttt{on(X,Y)}. While ILP based systems are usually able to solve variations of this environments, they rely on explicit representation of state and they cannot infer the state from the image. Here, we consider the task of stacking boxes on top of each other. We increase the difficulty of the problem compared to the previous examples \cite{dvzeroski2001relational,van2005survey,jiang2019neural} by considering the order of boxes and requiring that the stack is formed on top of the blue box (the blue box should be on the floor). To make sure the models learn generalization, we randomly place boxes on the floor in the beginning of each episode. We consider up to 5 boxes. Hence, the scene constants in our ILP setup is the set \texttt{\{a,b,c,d,e,floor\}}. The dimension of the observation images is 64x64x3 and no explicit relational information is available for the agents. The action space for the problem involving  $n$ boxes is $(n+1)\times (n+1)$ corresponding to all possibilities of moving a box (or the floor) on top of another box or the floor. Obviously some of the actions are not permitted, e.g., placing the floor on top of a box or moving a box that is already covered by another box. 

\begin{figure}
	\centering
	\includegraphics[width=0.380\textwidth]{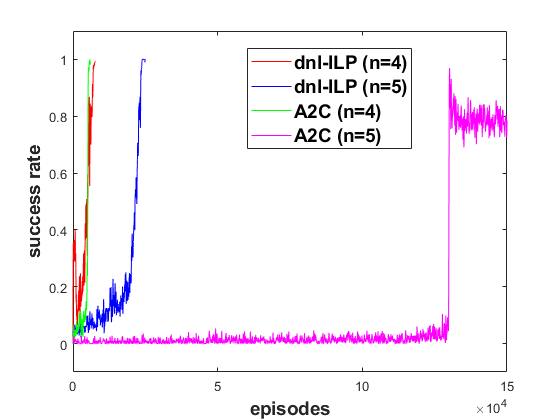}
	\vspace{-2mm}
	\caption{Comparing deep A2C and the proposed model on BoxWorld task}
	\label{fig:box_cmp}
\end{figure}

\begin{figure}
	\centering
	\includegraphics[width=0.380\textwidth]{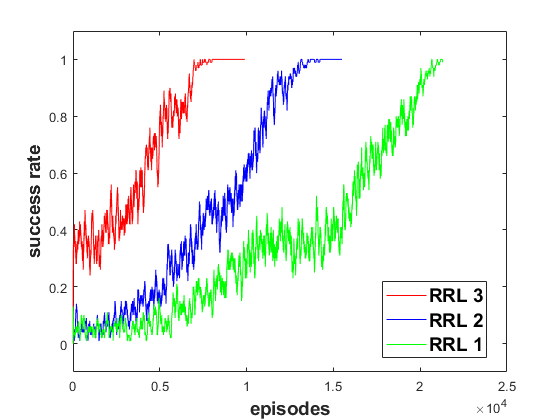}
	\vspace{-2mm}
	\caption{Effect of background knowledge on learning BoxWorld}
	\label{fig:box_bk}
\end{figure}

\paragraph{Comparing to Baseline:}
In the first experiment, we compare the performance of the proposed RRL technique to a baseline. For the baseline we consider standard deep A2C (with up to 10 agents) and we use the implementation in \texttt{stable-baseline} library \cite{stable-baselines}. We considered both MLP and CNN policies for the deep RL but we report the results for the CNN policy because of its superior performance. 
For the proposed RRL system, we use two convolutional layers with the kernel size of 3 and strides of 2 with $tanh$ activation function. We apply two layers of MLP with \texttt{softmax} activation functions to learn the groundings of the predicates \texttt{posH(X,Y)} and \texttt{posV(X,Y)}. Our presumed grid is $(n+1)\times(n+1)$ and we allow for positional constants  \texttt{\{0,1,\dots,n\}} to represent the locations in the grid in our ILP setting. 
As constraint we add penalty terms to make sure \texttt{posV(floor,0)} is True. We use vanilla gradient policy learning and to generate actions we define a learnable hypothesis predicate \texttt{move(X,Y)}. Since we have $n+1$ box constants (including floor), the groundings of this hypothesis correspond to the  $(n+1)\times(n+1)$ possible actions. Since the value of these groundings in dNL-ILP will be between 0 and 1, we generate \texttt{softmax} logits by multiplying these outputs by a large constant $c$ (e.g., $c=10$). For the  target predicate \texttt{move(X,Y)}, we allows for 6 rules in learning ( corresponding to dNL-DNF function with 6 disjunctions). The complete list of auxiliary predicates and parameters and weights used in the two models are given in appendix \ref{app:box}. As indicated in Fig. \ref{fig:ilp_rrl_state} and  defined in (\ref{eq:on}), we introduce predicate \texttt{on(X,Y)} as a function of the low-level state representation predicates \texttt{posV(X,Y)} and \texttt{posH(X,Y)}. We also introduce higher level concepts using these predicates to define the aboveness (i.e., \texttt{above(X,Y)}) as well as \texttt{isCovered(X,Y)}. 
Fig. \ref{fig:box_cmp} compares the average success per episode for the two models for the two cases of $n=4$ and $n=5$. The results shows that for the case of $n=4$, both models are able to learn a successful policy after around 7000 episodes. For the more difficult case of $n=5$, our proposed approach converges after around 20K episodes 
whereas  it takes more than 130K episodes for the A2C approach to converge, and even then it fluctuates and does not always succeed.
\paragraph{Effect of background knowledge:}
Contrary to the standard deep RL, in an RRL approach, we can introduce our prior knowledge into the problem via the powerful predicate language. By defining the structure of the problem via ILP, we can explicitly introduce inductive biases \cite{battaglia2018relational} which would restrict the possible form of the solution. We can speed up the learning process or shape the possible learnable actions even further by incorporating background knowledge. 
To examine the impact of the background knowledge on the speed of learning, 
we consider three cases for the BoxWorld problem involving $n=4$ boxes. The baseline model (RRL1) is as described before. In RRL2, we add another auxiliary predicate which defines the movable states as:

\begin{align*}
\text{movable(X,Y)} \leftarrow \neg \text{isCovered(X)}, \neg \text{isCovered(Y)},  \\\neg \text{same(A,B)}, \neg \text{isfloor(X)}, \neg \text{on(X,Y)}
\end{align*}
where $\neg$ indicates the negate of a term. In the third model (RRL3), we go one step further, and we force the target predicate \text{move(X,Y)} to incorporate the predicate \texttt{movable(X,Y)} in each of the conjunction terms. 
Fig. \ref{fig:box_bk} compares the learning performance of these models in terms of average success rate (between [0,1]) vs the number of episodes.
%
%
%
%
%
%
\paragraph{Interpretability:}
In the previous experiments, we did not consider the interpretability of the learned hypothesis. Since all the weights are fuzzy values, even though the learned hypothesis is still a parameterized symbolic function, it does not necessarily represent a valid Boolean formula. 
To achieve an interpretable result we add a small penalty as described in (\ref{eq:interpret}). We also add a few more state constraints to make sure the learned representation follow our presumed grid notations (see Appendix \ref{app:box} for details). The learned action predicate is found as: 
\begin{align*}
\text{move(X, Y)} &\leftarrow  \text{moveable(X, Y)},\, \neg \text{lower(X, Y)} \\
\text{move(X, Y)} &\leftarrow   \text{moveable(X, Y)},\, \text{isBlue(Y)} \\
\text{lower(X, Y)}  &\leftarrow  \text{posV(X, Z)},\, \text{posV(Y, T)},\, \text{lessthan(Z, T)}  
\end{align*}

\subsection{GridWorld Experiment}
\label{subsec:GridWorld}
\begin{figure}
	\centering
	\includegraphics[width=0.4\textwidth]{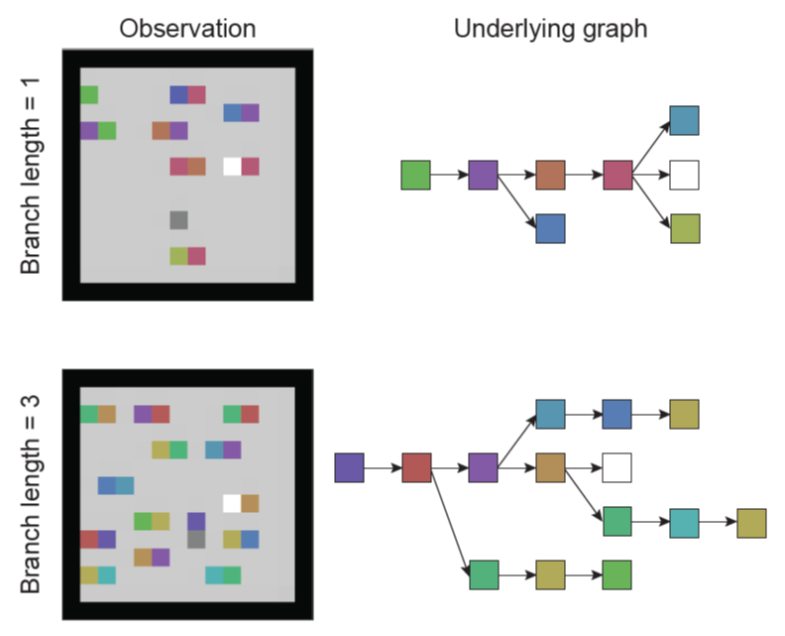}
	\caption{GridWorld environment \cite{zambaldi2018relational}} \label{fig:ilp_gridworld}
	\label{fig:ilp_boxworldaction}
\end{figure}
We use the GridWorld environment introduced in \cite{zambaldi2018relational} for this experiment. This environment is consisted of a $12\times12$ grid with keys and boxes randomly scattered. It also have an agent, represented by a single dark gray square box. The boxes are represented by two adjacent colors. The square on the right represents the box’s lock type whose color indicates which key can be used to open that lock. The square on the left indicates the content of the box which is inaccessible while the box is locked. The agent must collect the key before accessing the box.
%
When the agent has a key, provided that it walks over the lock box with the same color as its key, it can open the lock box, and then it must enter to the left box to acquire the new key which is inside the left box.
The agent cannot get the new key prior to successfully opening the lock box on the right side of the key box.
The goal is for the agent to open the gem box colored as white. We consider two difficulty levels. In the simple scenario, there is no (dead-end) branch. In the more difficult version, there can be one branch of dead end. An example of the environment and the branching scenarios is depicted in Fig.~\ref{fig:ilp_gridworld}.
This is a very difficult task involving complex reasoning. Indeed, in the original work it was shown that a multi agent A3C combined with a non-local learning attention model could only start to learn after processing $5\times10^8$ episodes. To make this problem easier to tackle, 
we modify the action space to include the location of any point in the grid instead of directional actions. Given this definition of the problem, the agent's task is to give the location of the next move inside the rectangular grid. Hence, the dimension of the action space is $144=12\times12$. 
For this environment, we define the predicates \texttt{color(X,Y,C)}, where $X,Y\in\{1,\dots12\}$, $C\in\{1,\dots,10\}$ and \texttt{hasKey(C)} to represent the state. Here, variables $X,Y$ denote the coordinates, and the variable $C$ is for the color. Similar to the BoxWorld game, we included a few auxiliary predicates such as \texttt{isBackground(X,Y)}, \texttt{isAgent(X,Y)} and \texttt{isGem(X,Y)} as part of the background knowledge. The representational power of ILP allows us to incorporate our prior knowledge about the problem into the model. As such we can include some higher level auxiliary helper predicates such as :
\begin{align*}
\text{isItem(X, Y)}&\leftarrow \ \neg \text{isBackground(X, Y)}, \neg \text{isAgent(X, Y)} \\
\text{locked(X, Y)}&\leftarrow \  \text{isItem(X, Y)}, \text{isItem(X,Z)}, \text{inc(Y, Z)}
\end{align*}
where predicate \texttt{inc(X,Y)} defines increments for integers (i.e., \texttt{inc(n,n+1)} is true for every integer n). 
The list of all auxiliary predicates used in this experiment as well as the parameters of the neural networks used in this experiment are given in Appendix \ref{app:grid}. 
Similar to previous experiments we consider two models, an A2C agent as the baseline and our proposed RRL model using the ILP language described in Appendix \ref{app:grid}.
\begin{table}[ht]
	\caption{Number of training episodes required for convergence }
	\label{tbl:results_block}
	\centering 
	\begin{tabular} { l  c  c c c c c c }
		\toprule
		model &   Without Branch & With Branch\\ 		
		\midrule
		proposed RRL & 700   & 4500  \\
		A2C &  $> 10^8$ & $> 10^8$ \\
		\bottomrule
	\end{tabular}
\end{table}
\begin{figure}
	\centering
	\includegraphics[width=0.35\textwidth]{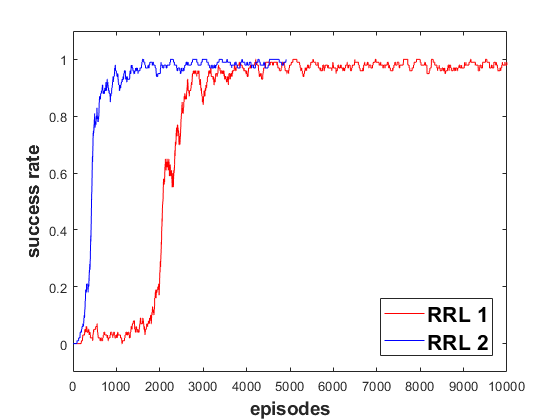}
	\vspace{-2mm}
	\caption{Effect of background knowledge on learning GridWorld}
	\label{fig:grid_bk}
\end{figure}
%
We listed the number of episodes it takes to converge in each setting in Table\ref{tbl:results_block}. As the results suggest, the proposed approach can learn the solution in both settings very fast. On the contrary, the standard deep A2C was not able to converge after $10^8$ episodes. 
This example restates the fact that incorporating our prior knowledge regarding the problem can significantly speed up the learning process.

Further, similar to the BoxWorld experiment, we study the importance of our background knowledge in the learning. In the first task (RRL1), we evaluate our model on the non-branching task by enforcing the action to include the $isItem(X,Y)$ predicate. In RRL2, we do not enforce this. As shown in Fig\ref{fig:grid_bk}, RRL1 model learns 4 times faster than RRL2. Arguably, this is because, enforcing the inclusion of  $isItem(X,Y)$ in the action hypothesis reduces the possibility of exploring irrelevant moves (i.e., moving to a location without any item).
%
%
\subsection{Relational Reasoning}
\label{subsec:SORTOFCLEVER}
Combining dNL-ILP with standard deep learning techniques is not limited to the RRL settings. In fact, the same approach can be used in other 
areas in which we wish to reason about the relations of objects.
To showcase this, we consider the relational reasoning task involving the Sort-of-CLEVR \cite{santoro2017simple} dataset. This dataset (See Fig.\ref{subfig:sortofclever}) consists of 2D images of some colored objects. The shape fo each object is either a rectangle or a circle and each image contains up to 6 objects.  The questions are hard-coded as fixed-length binary strings. Questions are either non-relational (e.g, "what is the color of the green object?") or relational (e.g., "what is the shape of the nearest object to the green object?"). In \cite{santoro2017simple}, the authors combined a CNN generated feature map with a special type of attention based non-local network in order to solve the problem. We use the same CNN network and similar to the GridWorld experiment, we learn the state representation using predicate \texttt{color(X,Y,C)} (the color of each cell in the grid) as well as \texttt{isCircle(X,Y)} which learn if the shape of an object is circle or not. Our proposed approach reaches the accuracy of 99\% on this dataset compared to the 94\% for the non-local approach presented in \cite{santoro2017simple}. The details of the model and the list of predicates in our ILP implementation is given in appendix \ref{app:relational}.


\section{Conclusion}
\label{sec:Conclusion}
In this paper, we proposed a novel deep Relational Reinforcement Learning (RRL) model based on a differentiable Inductive Logic Programming (ILP) that can effectively learn relational information from image. We showed how this model can take the expert background knowledge and incorporate it into the learning problem using appropriate predicates. The differentiable ILP allows an end to end optimization of the entire framework for learning the policy in RRL. We showed the performance of the proposed RRL framework using environments such as BoxWorld and GridWorld.


\nocite{langley00}

\bibliography{refs_payani,references-ILP}

\begin{thebibliography}{22}
\providecommand{\natexlab}[1]{#1}
\providecommand{\url}[1]{\texttt{#1}}
\expandafter\ifx\csname urlstyle\endcsname\relax
  \providecommand{\doi}[1]{doi: #1}\else
  \providecommand{\doi}{doi: \begingroup \urlstyle{rm}\Url}\fi

\bibitem[Abadi et~al.(2016)Abadi, Barham, Chen, Chen, Davis, Dean, Devin,
  Ghemawat, Irving, Isard, et~al.]{abadi2016tensorflow}
Abadi, M., Barham, P., Chen, J., Chen, Z., Davis, A., Dean, J., Devin, M.,
  Ghemawat, S., Irving, G., Isard, M., et~al.
\newblock Tensorflow: A system for large-scale machine learning.
\newblock In \emph{12th $\{$USENIX$\}$ Symposium on Operating Systems Design
  and Implementation ($\{$OSDI$\}$ 16)}, pp.\  265--283, 2016.

\bibitem[Allamanis et~al.(2017)Allamanis, Brockschmidt, and
  Khademi]{allamanis2017learning}
Allamanis, M., Brockschmidt, M., and Khademi, M.
\newblock Learning to represent programs with graphs.
\newblock \emph{arXiv preprint arXiv:1711.00740}, 2017.

\bibitem[Battaglia et~al.(2018)Battaglia, Hamrick, Bapst, Sanchez-Gonzalez,
  Zambaldi, Malinowski, Tacchetti, Raposo, Santoro, Faulkner,
  et~al.]{battaglia2018relational}
Battaglia, P.~W., Hamrick, J.~B., Bapst, V., Sanchez-Gonzalez, A., Zambaldi,
  V., Malinowski, M., Tacchetti, A., Raposo, D., Santoro, A., Faulkner, R.,
  et~al.
\newblock Relational inductive biases, deep learning, and graph networks.
\newblock \emph{arXiv preprint arXiv:1806.01261}, 2018.

\bibitem[Blockeel \& De~Raedt(1998)Blockeel and De~Raedt]{blockeel1998top}
Blockeel, H. and De~Raedt, L.
\newblock Top-down induction of first-order logical decision trees.
\newblock \emph{Artificial intelligence}, 101\penalty0 (1-2):\penalty0
  285--297, 1998.

\bibitem[Bryant et~al.(1999)Bryant, Muggleton, Page, Sternberg,
  et~al.]{bryant1999combining}
Bryant, C., Muggleton, S., Page, C., Sternberg, M., et~al.
\newblock Combining active learning with inductive logic programming to close
  the loop in machine learning.
\newblock In \emph{AISB’99 Symposium on AI and Scientific Creativity}, pp.\
  59--64. Citeseer, 1999.

\bibitem[D{\v{z}}eroski et~al.(1998)D{\v{z}}eroski, De~Raedt, and
  Blockeel]{dvzeroski1998relational}
D{\v{z}}eroski, S., De~Raedt, L., and Blockeel, H.
\newblock Relational reinforcement learning.
\newblock In \emph{International Conference on Inductive Logic Programming},
  pp.\  11--22. Springer, 1998.

\bibitem[D{\v{z}}eroski et~al.(2001)D{\v{z}}eroski, De~Raedt, and
  Driessens]{dvzeroski2001relational}
D{\v{z}}eroski, S., De~Raedt, L., and Driessens, K.
\newblock Relational reinforcement learning.
\newblock \emph{Machine learning}, 43\penalty0 (1-2):\penalty0 7--52, 2001.

\bibitem[Evans \& Grefenstette(2018)Evans and Grefenstette]{evans2018learning}
Evans, R. and Grefenstette, E.
\newblock Learning explanatory rules from noisy data.
\newblock \emph{Journal of Artificial Intelligence Research}, 61:\penalty0
  1--64, 2018.

\bibitem[Hessel et~al.(2017)Hessel, Modayil, van Hasselt, Schaul, Ostrovski,
  Dabney, Horgan, Piot, Azar, and Silver]{hessel2017rainbow}
Hessel, M., Modayil, J., van Hasselt, H., Schaul, T., Ostrovski, G., Dabney,
  W., Horgan, D., Piot, B., Azar, M., and Silver, D.
\newblock Rainbow: Combining improvements in deep reinforcement learning, 2017.

\bibitem[Hill et~al.(2018)Hill, Raffin, Ernestus, Gleave, Kanervisto, Traore,
  Dhariwal, Hesse, Klimov, Nichol, Plappert, Radford, Schulman, Sidor, and
  Wu]{stable-baselines}
Hill, A., Raffin, A., Ernestus, M., Gleave, A., Kanervisto, A., Traore, R.,
  Dhariwal, P., Hesse, C., Klimov, O., Nichol, A., Plappert, M., Radford, A.,
  Schulman, J., Sidor, S., and Wu, Y.
\newblock Stable baselines.
\newblock \url{https://github.com/hill-a/stable-baselines}, 2018.

\bibitem[Jiang \& Luo(2019)Jiang and Luo]{jiang2019neural}
Jiang, Z. and Luo, S.
\newblock Neural logic reinforcement learning, 2019.

\bibitem[Mnih et~al.(2013)Mnih, Kavukcuoglu, Silver, Graves, Antonoglou,
  Wierstra, and Riedmiller]{mnih2013playing}
Mnih, V., Kavukcuoglu, K., Silver, D., Graves, A., Antonoglou, I., Wierstra,
  D., and Riedmiller, M.
\newblock Playing atari with deep reinforcement learning.
\newblock \emph{arXiv preprint arXiv:1312.5602}, 2013.

\bibitem[Mnih et~al.(2016)Mnih, Badia, Mirza, Graves, Lillicrap, Harley,
  Silver, and Kavukcuoglu]{mnih2016asynchronous}
Mnih, V., Badia, A.~P., Mirza, M., Graves, A., Lillicrap, T., Harley, T.,
  Silver, D., and Kavukcuoglu, K.
\newblock Asynchronous methods for deep reinforcement learning.
\newblock In \emph{International conference on machine learning}, pp.\
  1928--1937, 2016.

\bibitem[Narayanan et~al.(2017)Narayanan, Chandramohan, Venkatesan, Chen, Liu,
  and Jaiswal]{narayanan2017graph2vec}
Narayanan, A., Chandramohan, M., Venkatesan, R., Chen, L., Liu, Y., and
  Jaiswal, S.
\newblock graph2vec: Learning distributed representations of graphs.
\newblock \emph{arXiv preprint arXiv:1707.05005}, 2017.

\bibitem[Payani \& Fekri(2018)Payani and Fekri]{payani2018}
Payani, A. and Fekri, F.
\newblock Decoding ldpc codes on binary erasure channels using deep recurrent
  neural-logic layers.
\newblock In \emph{Turbo Codes and Iterative Information Processing (ISTC),
  2018 International Symposium On}. IEEE, 2018.

\bibitem[Payani \& Fekri(2019)Payani and Fekri]{payani2019Learning}
Payani, A. and Fekri, F.
\newblock Inductive logic programming via differentiable deep neural logic
  networks.
\newblock \emph{arXiv preprint arXiv:1906.03523}, 2019.

\bibitem[Santoro et~al.(2017)Santoro, Raposo, Barrett, Malinowski, Pascanu,
  Battaglia, and Lillicrap]{santoro2017simple}
Santoro, A., Raposo, D., Barrett, D.~G., Malinowski, M., Pascanu, R.,
  Battaglia, P., and Lillicrap, T.
\newblock A simple neural network module for relational reasoning.
\newblock In \emph{Advances in neural information processing systems}, pp.\
  4967--4976, 2017.

\bibitem[Van~Hasselt et~al.(2016)Van~Hasselt, Guez, and Silver]{van2016deep}
Van~Hasselt, H., Guez, A., and Silver, D.
\newblock Deep reinforcement learning with double q-learning.
\newblock In \emph{Thirtieth AAAI Conference on Artificial Intelligence}, 2016.

\bibitem[Van~Otterlo(2005)]{van2005survey}
Van~Otterlo, M.
\newblock A survey of reinforcement learning in relational domains.
\newblock \emph{Centre for Telematics and Information Technology (CTIT)
  University of Twente, Tech. Rep}, 2005.

\bibitem[Vaswani et~al.(2017)Vaswani, Shazeer, Parmar, Uszkoreit, Jones, Gomez,
  Kaiser, and Polosukhin]{vaswani2017attention}
Vaswani, A., Shazeer, N., Parmar, N., Uszkoreit, J., Jones, L., Gomez, A.~N.,
  Kaiser, {\L}., and Polosukhin, I.
\newblock Attention is all you need.
\newblock In \emph{Advances in neural information processing systems}, pp.\
  5998--6008, 2017.

\bibitem[Watkins \& Dayan(1992)Watkins and Dayan]{watkins1992q}
Watkins, C.~J. and Dayan, P.
\newblock Q-learning.
\newblock \emph{Machine learning}, 8\penalty0 (3-4):\penalty0 279--292, 1992.

\bibitem[Zambaldi et~al.(2018)Zambaldi, Raposo, Santoro, Bapst, Li, Babuschkin,
  Tuyls, Reichert, Lillicrap, Lockhart, et~al.]{zambaldi2018relational}
Zambaldi, V., Raposo, D., Santoro, A., Bapst, V., Li, Y., Babuschkin, I.,
  Tuyls, K., Reichert, D., Lillicrap, T., Lockhart, E., et~al.
\newblock Relational deep reinforcement learning.
\newblock \emph{arXiv preprint arXiv:1806.01830}, 2018.

\end{thebibliography}
\bibliographystyle{icml2020}

\onecolumn
\newpage
\appendix
\section{BoxWorld Experiment}
\label{app:box}
For the problem consists of $n$ box, we need $n+1$ constants of type \texttt{box} (note that we consider the floor as one of the boxes in our problem definition). Additionally, we define numerical constants $\{0,\dots,n\}$ to represent box coordinates using in posH and posV predicates. For the numerical constants we define orderings via extensional predicates \texttt{lt}/2 and \texttt{inc}/2. For the box constants, we define two extensional predicates \texttt{same/2} and \texttt{isBlue/1}. Here, by \texttt{p}/$N$ we mean predicate \texttt{p} of arity $N$ (i.e., \texttt{p} has $N$ arguments). Since these are extensional predicates, their truth values are fixed in the beginning of the program via the background facts. For example, for predicate \texttt{inc}/2 which defines the increment by one for the natural numbers, we need to set these background facts the beginning: \texttt{\{inc(0,1), inc(1,2),\dots,inc(n-1,n)\}}. Similarly, for the predicate \texttt{lt} (short for lessThan) this set includes items such as \texttt{\{lt(0,1), lt(0,2),\dots,lt(n-1,n)\}}. 
It is worth noting that introducing predicates such as \texttt{isBlue} for boxes does not mean we already know which box is the blue one (the target box that needs to be first box in the stack). This predicate merely provides a way for the system to distinguish between boxes. Since in our learned predicates we can only use symbolic atoms (predicates with variables) and in dNL-ILP implementation, no constant is allowed in forming the hypothesis, this method allows for referring to an specific constant in the learned action predicates. Here, for example, we make an assumption that box \texttt{a} in our list of constants corresponds to the blue box via the background fact \texttt{isBlue(a)}. Table \ref{tbl:ilpBoxWorld} explains all extensional predicates as well as those helper auxiliary predicates that was used in our BoxWorld program. 

So far in our discussions, we have not  distinguished between the type of variables. However, in the dNL-ILP implementation, the type of each variable should be specified in the signature of each defined predicate. This allows the dNL-ILP to use only valid combination of variables to generate the symbolic formulas. In the BoxWorld experiment, we have two types of constants $T_b$ and $T_p$ referring to the box and numeric constants,  respectively. In Table\ref{tbl:ilpBoxWorld}, for each defined predicate, the list of variables and their corresponding types are given, where for example X($T_p$) states that the type of variable X is $T_p$.

For clarity, Table \ref{tbl:ilpBoxWorld} is divided into two sections. In the top section the constants are defined. In the other section, 4 groups of predicates are presented: (i) state representation predicates that their groundings are learned from image, (ii) extensional predicates which are solely defined by background facts, (iii) auxiliary predicates and, (iv) the signature for the target predicate that is used to represent the actions in the policy gradient scheme. To learn the policy, we used discount factor of 0.7, and we use ADAM optimizer with learning rate of 0.002. We set the maximum number of steps for each episode to 20. To learn the feature map, we used two layers of CNNs with kernel size of 5 and strides of 3 and we applied  fully connected layers with \texttt{softmax} activation functions to learn the groundings of the state representation predicates \texttt{posH} and \texttt{posV}.

\section{GridWorld Experiment}
\label{app:grid}
\begin{figure}
	\centering
	\includegraphics[width=0.5\textwidth]{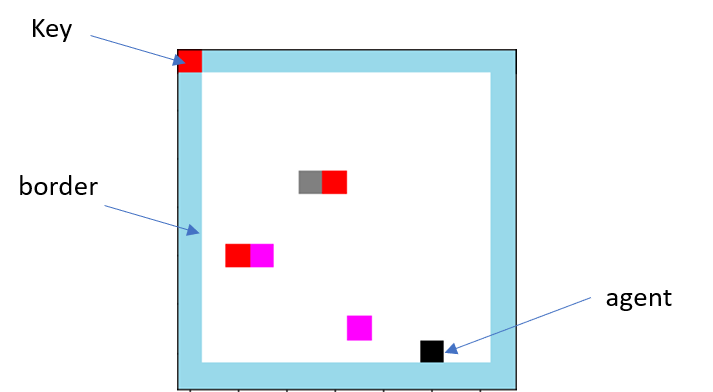}
	\caption{An example GridWorld scene}
	\label{fig:grid}
	\vspace{-.16in}
\end{figure}

In the GridWorld experiment, we distinguish between constants used for vertical and horizontal coordinates as shown Table \ref{tbl:ilpGridWorld}. This reduces the number of possible symbolic atoms and makes the convergence faster.
We also define 10 constants of type $T_c$ to represent each of the 10 possible colors for the grid cells in the scene. The complete list of all the extensional and auxiliary predicates for this program is listed in Table \ref{tbl:ilpGridWorld}. In this task, to incorporate our knowledge about the problem, we define the concept of \texttt{item} via predicate \texttt{isItem(X,Y)}. Using this definition, an \texttt{item} represents any of the the cells that are neither background nor agent. By incorporating this concept, we can further define higher level concepts via \texttt{locked} and \texttt{isLock} predicates which define the two adjacent items as a locked item and its corresponding key, respectively. The dimension of the observation image is 112x112x3. We consider two types of networks to extract grid colors from the image. In the first method, we apply three layers of convolutional network with kernel sizes of $[5,3,3]$ and strides of 2 for all. We use \texttt{relu} activation function apply batch normalization after each layer. The number of features in each layer are set to 24. We take the feature map of size $14\times14\times24$ and apply two layers MLP with dimensions of 64 and 10 (10 is the the number of color constants) and activation functions of \texttt{relu} and \texttt{softmax}, to generate the grounding matrix $G$ (the dimension of $G$ is 14x14x10). 
As can be seen in Fig \ref{fig:grid}, the current key is located at the top left border. We extract the grounding for predicate \texttt{hasKey} from the value of $G[0,0,:]$ and the groundings of the predicate \texttt{color} by discarding the border elements from $G$ (i.e., $G[1..12,1..12,:]$). 

Alternatively, and because of the simplicity of the environment, instead of using CNNs, we may take the center of each box to directly create the feature map of size $14\times14\times3$ and then apply the MLP to generate the groundings. In our experiments we tested both scenarios. Using the CNN approach the speed of convergence in our method was around 2 times slower. For the A2C algorithm, we could not solve the problem using either of these approaches. We set the maximum number of steps in an episode to 50 and we use learning rate of .001. For our models, we use discount factor of 0.9 and for the A2C we tested various numbers in range of $0.9$ to $0.99$ to find the best solution. 

\section{Relational Reasoning Experiment}
\label{app:relational}
This task was introduced as a benchmark for learning relational information from images in \cite{santoro2017simple}. In this task, the objects of two types (circle or rectangle) and in 6 different colors are randomly placed in a  4x4 grid and questions like \emph{'is the blue object circle'} or \emph{'what is the color of nearest object to blue'} should be answered. To formulate this problem via ILP, instead of using grid coordinates (vertical and horizontal positions as in past experiments) we consider a flat index as can be seen in Table \ref{tbl:flatGrid}. As such, we will have 16 positional constants of type $T_p$ to represent grid positions as well as 6 constants of type $T_c$ to represent the color of each object. In the original problem in \cite{santoro2017simple}, a question is represented as binary vector of length 11. The first 6 bits represent the one-hot representation of the color of the object in question. The next 5 bits represent one of the five possible types of questions, i.e., \emph{'is it a circle or a rectangle'} and \emph{'the color of the nearest object?'} for example. 

We define a nullary predicate for each of 11 possible bits in the question via predicates \texttt{isQ0()},\dots,\texttt{isQ10()}. Similar to to the original paper we use 4 layers of CNNs with \texttt{relu} activation function and batch normalization to obtain a feature map of size $4x4X24$. We use kernel size of 5 and strides of [3,3,2,2] for each of the layers. By applying fully connected layers to the feature map, we learn the groundings of predicates \texttt{color(X,Y,Z)}, \texttt{isObject(X,Y)} and \texttt{isCircle(X,Y)}. We define some auxiliary predicates as shown in Table \ref{tbl:ilpRelational}. For each of the 10 possible answers, we create one nullary predicate. The vector of size 10 that is created by the the groundings of these 10 predicates (i.e., \texttt{isAnswer0()}, \dots) are then used to calculate the cross entropy loss between the network output and the provided ground truth. We need to mention that in the definition of the auxiliary predicate \texttt{qa(X)}, we exploit our prior knowledge regarding the fact that the first 6 bits of the question vector correspond to the colors of the object. Without introducing this predicate, the convergence of our model is significantly slower. For example, while by incorporating this predicate we need around 30000 training samples for convergence, it would take than 200000 training samples without this predicate.

\begin{table}[]
	\caption{Flat index for a grid of 4 by 4 used in relational learning task}
	\vspace{3mm}
	\label{tbl:flatGrid}
	\centering 
	\begin{tabular}{|l|l|l|l|}
		\hline
		0  & 1  & 2  & 3  \\ \hline
		4  & 5  & 6  & 7  \\ \hline
		8  & 9  & 10 & 11 \\ \hline
		12 & 13 & 14 & 15 \\ \hline
	\end{tabular}
\end{table}

\section{Asterix Experiment}
\label{app:asterix}
In Asterix game (an Atari 2600 game), the player controls a unit called Asterix to collect as many objects as possible while avoiding deadly predators. Even though it is a rather simple game, many standard RL algorithms score significantly below the human level. for example, as reported in \cite{hessel2017rainbow}, DQN and A3C achieve the scores of 3170 and 22140 on average, respectively even after processing hundreds of millions of frames. Here, our goal is not to outperform the current approaches. Instead, we demonstrate that by using a very simple language for describing the problem, an agent can achieve scores in the range of 30K-40K with only a few thousands of training scenes. For this game we consider the active part of the scene as a 8x12 grid. For simplicity, we consider only 4 types of objects; agent, left to right moving predator, right to left moving predator and finally the food objects. The dimension of the input image is 128x144. We use 4 convolutional layers with strides of [(2,3),(2,1),(2,2),(2,2)] and kernel size of 5 to generate a feature map of size 8x12x48. By applying 4 fully connected layers with \texttt{softmax} activation function we learn the groundings of the predicates corresponding to the four types of objects, i.e., \texttt{O1(X,Y)},\dots,\texttt{O4(X,Y)}. The complete list of predicates for this experiment is listed in Table \ref{tbl:ilpAsterix}. We learn 5 nullary predicates corresponding to the 5 direction of moves (i.e., no move, left,right,up,down) and use the same policy gradient learning approach as before. The notable auxiliary predicates in the chosen language are the 4 helper predicates that define bad moves. For example, \texttt{badMoveUp()} states that an upward move is bad when there is a predator in the close neighborhood of the agent and in the top row. Similarly, \texttt{badMoveLeft()} is defined to state that when a predator is coming from left side of an agent, it is a bad idea to move left. 

\begin{figure}
	\centering
	\includegraphics[width=0.5\textwidth]{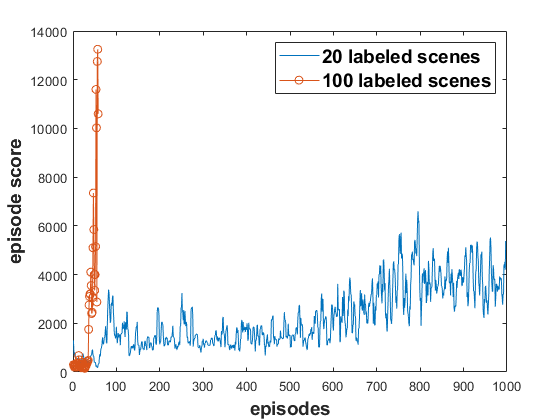}
	\caption{Score during training in Astreix experiment}
	\label{fig:asterixplot}
	\vspace{-.16in}
\end{figure}

However, given the complexity of the scene and the existence of some overlappings between the objects, learning the representation of the state is not as easy as the previously explored experiments. To help the agent learn the presumed grid presentation, we provide a set of labeled scene (a semi-supervised approach) and we penalize the objective function using the loss that is calculated between these labels and the learned representation. Fig. \ref{fig:asterixplot} shows the learning curve for two cases of the using 20 and 50 randomly generated labeled scenes. In the case of 50  provided labels, the agent finally learns to score around 30-40K each episodes. Please note that we did not include all the possible objects that are encountered during later stages of the game and we use a simplistic representation and language just to to demonstrate the application of RRL framework in more complex scenarios.

\begin{table}[ht]
	\caption{ILP definition of the BoxWorld}
	\label{tbl:ilpBoxWorld}
	\centering 
	
	\begin{tabular}{ l  l  l}
		\toprule
		Constants & Description & Values \\ \hline
		$T_b$ & box constants & \{a, b, c, d, floor\} \\
		$T_p$ & coordinate position constants & \{0, 1, 2, \dots, n\} \\ \\
		
	\end{tabular}
	\begin{tabular} { l | l | l }
		\toprule
		Predicate &   Variables & Definition\\ 		
		\midrule
		posH(X,Y)& X($T_b$), Y($T_p$) & learned from Image  \\
		posV(X,Y)& X($T_b$), Y($T_p$) & learned from Image  \\ \hline \\
		isFloor(X)& X($T_b$)&  isFloor(floor) \\
		isBlue(X)& X($T_b$)&  isBlue(a) \\
		isV1(X)& X($T_p$)&  isV1(1)\\
		inc(X,Y)& X($T_p$), Y($T_p$) & inc(0,1), inc(1,2), \dots, inc(n-1,n)\\
		lt(X,Y)& X($T_p$), Y($T_p$) & lt(0,1), lt(0,2), \dots, lt(n-1,n)\\ \hline\\
		same(X,Y)&X($T_b$), Y($T_b$) & same(a,a), same(b,b), \dots, same(floor,floor)\\
		sameH(X,Y)&X($T_b$), Y($T_b$), Z($T_p$) & posH(X,Z), posH(Y,Z)\\
		sameV(X,Y)&X($T_b$), Y($T_b$), Z($T_p$) & posV(X,Z), posV(Y,Z)\\
		above(X,Y)&X($T_b$), Y($T_b$), Z($T_p$), T($T_p$) & sameH(X,Y), posV(X,Z), posV(Y,T), lt(T,Z)\\
		below(X,Y)&X($T_b$), Y($T_b$), Z($T_p$), T($T_p$) & sameH(X,Y), posV(X,Z), posV(Y,T), lt(Z,T)\\
		on(X,Y)& X($T_b$), Y($T_b$), Z($T_p$),T($T_p$)  &  sameH(X,Y), posV(X,Z), posV(Y,T), inc(T,Z)\\
		isCovered(X)& X($T_b$), Y($T_b$) & On(Y,X), $\neg$isFloor(X)\\
		moveable(X,Y)& X($T_b$), Y($T_b$) & $ \neg \text{isCovered(X)}, \neg \text{isCovered(Y)},  \neg \text{same(A,B)}, \neg \text{isfloor(X)}, \neg \text{on(X,Y)}$ \\ \hline\\
		move(X,Y)&X($T_b$), Y($T_b$), Z($T_p$),T($T_p$) &  Action predicate that is learned via policy gradient\\  
		& & \\
		\bottomrule
	\end{tabular}
\end{table}
\begin{table}[ht]
	\caption{ILP definition of the GridWorld}
	\label{tbl:ilpGridWorld}
	\centering 
	
	\begin{tabular}{ l   l   l}
		\toprule
		Constants & Description & Values \\ \hline
		$T_v$ & vertical coordinates & \{0,1,2,\dots,11\} \\
		$T_h$ & horizontal coordinates & \{0,1,2,\dots,11\} \\
		$T_c$ & cell color code & \{0,1,2,\dots,9\} \\ \\
		
	\end{tabular}
	\begin{tabular} { l | l | l }
		\toprule
		Predicate &   Variables & Definition\\ 		
		\midrule
		color(X,Y,Z)& X($T_v$), Y($T_h$), Z($T_c$) & learned from Image  \\ 
		hasKey(X)& X($T_c$) & Learned from Image  \\ 
		\hline 
		
		\\
		incH(X,Y)& X($T_h$)& incH(0,1),\dots,incH(10,11)  \\
		isC0(X)& X($T_c$)&  isC0(0) \\
		isC1(X)& X($T_c$)&  isC1(1) \\
		isC2(X)& X($T_c$)&  isC2(2) \\ \hline \\
		isBK(X,Y)& X($T_v$), X($T_h$), Z($T_c$)&  color(X,Y,Z), isC0(Z) \\
		isAgent(X,Y)& X($T_v$), X($T_h$), Z($T_c$)&  color(X,Y,Z), isC1(Z) \\
		isGem(X,Y)& X($T_v$), X($T_h$), Z($T_c$)&  color(X,Y,Z), isC2(Z) \\
		isItem(X,Y)& X($T_v$), X($T_h$)&  $\neg$ isBK(X,Y), $\neg$ isAgent(X,Y),  \\
		locked(X)& X($T_v$), X($T_h$), Z($T_h$)&  isItem(X,Y), isItem(X,Z), incH(Y,Z)\\
		isLock(X)& X($T_v$), X($T_h$), Z($T_h$)&  isItem(X,Y), isItem(X,Z), incH(Z,Y)\\ \hline \\
		
		move(X,Y)&X($T_v$), Y($T_h$), Z($T_c$) &  Action predicate that is learned via policy gradient\\  
		& & \\
		\bottomrule
	\end{tabular}
\end{table}
\begin{table}[ht]
	\caption{ILP definition of the relational reasoning task}
	\label{tbl:ilpRelational}
	\centering 
	
	\begin{tabular}{ l   l   l}
		\toprule
		Constants & Description & Values \\ \hline
		$T_p$ & Flat position of items in a 4 by 4 grid & \{0,1,2,\dots,15\} \\
		$T_c$ & color of an item & \{0,1,2,\dots,5\} \\
		
	\end{tabular}
	\begin{tabular} { l | l | l }
		\toprule
		Predicate &   Variables & Definition\\ 		
		\midrule
		isQ0()&  & Given as a binary value\\ 
		\dots&  & \dots\\ 
		isQ10()&  & Given as a binary value\\ 
		color(X,Y) & X($T_p$), Y($T_c$) & Learned from Image  \\ 
		isCircle(X,Y) & X($T_p$)& Learned from Image  \\ 
		isObject(X,Y) & X($T_p$)& Learned from Image  \\ 
		\hline 
		\\
		equal(X,Y)& X($T_p$), Y($T_p$) & equal(0,0),\dots,incH(15,15)  \\
		lt(X,Y)& X($T_p$), Y($T_p$), Z($T_p$) &  \begin{tabular}{l}true if distance between grid cells corresponding to X and Y \\is less than distance between X and Z (see Table \ref{tbl:flatGrid})\\ \end{tabular}  \\
		left(X)& X($T_p$)&  left(0),left(1),\dots,left(12),left(13) \\
		right(X)& X($T_p$)&  right(2),right(3),\dots,right(14),right(15) \\
		top(X)& X($T_p$)&  top(0),left(1),\dots,left(6),left(7) \\
		bottom(X)& X($T_p$)&  bottom(8),left(9),\dots,left(14),left(15) \\ \hline
		\\
		closer(X,Y,Z)& X($T_p$), X($T_p$), Z($T_p$)&  isObject(X), isObject(Y), isObject(Z), lt(X,Y,Z) \\
		farther(X,Y,Z)& X($T_p$), X($T_p$), Z($T_p$)&  isObject(X), isObject(Y), isObject(Z), gt(X,Y,Z) \\
		notClosest(X,Y)& X($T_p$), X($T_p$), Z($T_p$)&  closer(X,Z,Y) \\
		notFarthest(X,Y)& X($T_p$), X($T_p$), Z($T_p$)&  farther(X,Z,Y) \\ \hline 
		
		qa(X)& X($T_p$), Y($T_c$)&   
		\begin{tabular}{l}
			isQ0(), color(X,Y), isC0(Y)\\
			isQ1(), color(X,Y), isC1(Y)\\
			isQ2(), color(X,Y), isC2(Y)\\
			isQ3(), color(X,Y), isC3(Y)\\
			isQ4(), color(X,Y), isC4(Y)\\
			isQ5(), color(X,Y), isC5(Y)\\
		\end{tabular}
		\\ \hline 
		
		\\
		isAnswer0()&X($T_p$), Y($T_p$) &  The learned hypothesis : Is answer is 0\\  
		\dots & \dots & \dots \\
		isAnswer9()&X($T_p$), Y($T_p$) &  The learned hypothesis : Is answer is 9\\  
		& & \\
		\bottomrule
	\end{tabular}
\end{table}

\begin{table}[ht]
	\caption{ILP definition of the Asterix experiment}
	\label{tbl:ilpAsterix}
	\centering 
	
	\begin{tabular}{ l   l   l}
		\toprule
		Constants & Description & Values \\ \hline
		$T_v$ & vertical coordinates & \{0,1,2,\dots,8\} \\
		$T_h$ & horizontal coordinates & \{0,1,2,\dots,12\} \\
		
	\end{tabular}
	\begin{tabular} { l | l | l }
		\toprule
		Predicate &   Variables & Definition\\ 		
		\midrule
		O1(X,Y)& X($T_v$), Y($T_h$),  & learned from Image : objects of type agent \\ 
		O2(X,Y)& X($T_v$), Y($T_h$),  & learned from Image : objects of type L2R predator \\ 
		O3(X,Y)& X($T_v$), Y($T_h$),  & learned from Image : objects of type R2L predator  \\ 
		O4(X,Y)& X($T_v$), Y($T_h$),  & learned from Image : objects of type food \\ 
		\hline 
		
		\\
		isV0(X)& X($T_v$)& isV(0)  \\
		isV11(X)& X($T_v$)& isV11(11)  \\
		isH0(X)& X($T_h$)& isH0(0)  \\
		isH7(X)& X($T_h$)& isH7(7)  \\
		incV(X,Y)& X($T_v$), X($T_v$)& incV(0,1), \dots,incV(6,7)\\
		ltH(X,Y)& X($T_h$), X($T_h$)& ltH(0,1),ltH(0,2), \dots,ltH(11,12)\\
		closeH(X,Y) & X($T_h$), Y($T_h$) & true if $|X-Y| \le 2$ \\
		\hline \\
		
		agentH(X)& X($T_v$), Y($T_h$)&  O1(X,X)\\
		agentV(X)& X($T_h$), Y($T_v$)&  O1(Y,X)\\
		predator(X,Y) & X($T_v$), Y($T_h$) &  \begin{tabular}{l} O2(X,X) \\ O3(X,Y) \\ \end{tabular} \\
		agent() & X($T_v$)&  agentV(X) \\
		badMoveUp() & X($T_v$), Y($T_h$), Z($T_v$), T($T_h$)        &  O1(X,Y), predator(Z,T), incV(Z,X), closeH(Y,T) \\
		badMoveDown() & X($T_v$), Y($T_h$), Z($T_v$), T($T_h$)        &  O1(X,Y), predator(Z,T), incV(X,Z), closeH(Y,T) \\
		badMoveLeft() & X($T_v$), Y($T_h$), Z($T_h$)     &  O1(X,Y), O2(X,Z), ltH(Z,Y), closeH(Z,Y)\\
		badMoveRight() & X($T_v$), Y($T_h$), Z($T_h$)     &  O1(X,Y), O3(X,Z), ltH(Y,Z), closeH(Z,Y)\\
		\\ \hline \\
		moveUp() & X($T_v$), Y($T_h$), Z($T_v$), T($T_h$),  &  Action predicate that is learned via policy gradient\\  
		moveDown() & X($T_v$), Y($T_h$), Z($T_v$), T($T_h$),  &  Action predicate that is learned via policy gradient\\  
		moveLeft() & X($T_v$), Y($T_h$), Z($T_v$), T($T_h$),  &  Action predicate that is learned via policy gradient\\  
		moveRight() & X($T_v$), Y($T_h$), Z($T_v$), T($T_h$),  &  Action predicate that is learned via policy gradient\\  
		moveNOOP() & X($T_v$), Y($T_h$), Z($T_v$), T($T_h$),  &  Action predicate that is learned via policy gradient\\  
		& & \\
		\bottomrule
	\end{tabular}
\end{table}

\end{document}